\documentclass[journal,10pt]{IEEEtran}
\usepackage[T1]{fontenc}
\usepackage{amsmath,amssymb}
\usepackage{graphicx}
\usepackage[nocompress]{cite}
\usepackage[obeyspaces,spaces,hyphens]{url}
\usepackage[hidelinks]{hyperref}

\AtBeginDocument{\lefthyphenmin=2\righthyphenmin=3}
\newcommand{\method}{SpriteStage}
\newcommand{\state}{\mathcal{S}}
\newcommand{\program}{\mathcal{P}}
\newcommand{\evidence}{\mathcal{E}}

\begin{document}
\bstctlcite{IEEEcontrol}

\title{SpriteStage: Long Horizon Animation Generation via Progressive Scene Programming and Replay}

\author{Zhaoqi Wang, Zijian Zhang\textsuperscript{*}, Jiamou Liu%
\thanks{Zhaoqi Wang and Zijian Zhang are with the School of Cyberspace Science and Technology, Beijing Institute of Technology, Beijing 100081, China (e-mail: \{wang\_zhaoqi, zhangzijian\}@bit.edu.cn).}%
\thanks{Jiamou Liu is with the School of Computer Science, The University of Auckland, Auckland, New Zealand (e-mail: \mbox{jiamou.liu@auckland.ac.nz}).}%

\thanks{\textsuperscript{*}Corresponding author: Zijian Zhang.}}

\maketitle

\begin{abstract}
Video generation models synthesize visual sequences from text and image inputs. Extending video generation to long animations requires actions, dialogue, camera coverage, and narrative developments to remain coordinated across scenes. Executable scene programs make these decisions available for inspection. Constructing an entire story before observing its execution can leave later scenes dependent on unrealized events. We present SpriteStage, a progressive scene programming framework that builds animation through iterative enactment and engine reorchestration. Iterative enactment generates each scene from an accepted world checkpoint and narrative snapshot, executes it on a shared spatial stage, and uses runtime and visual feedback to revise the relevant actions or story beats. The accepted outcome becomes the starting state for the next scene. Engine reorchestration maps the resulting draft to production assets and engine capabilities while retaining its events, geometry, and timing. Inspection of the assembled sequence and frozen replay carry the accepted performance into the final video. Three complete animations, ranging from approximately 9.9 to 14.3 minutes, demonstrate the framework's effectiveness in producing extended narratives through scene programming and engine execution.
\end{abstract}


\section{Introduction}
Video generation models such as Seedance~1.0, Wan, and HunyuanVideo have advanced visual quality, motion synthesis, and control from text prompts~\cite{seedance1,wan,hunyuanvideo}. Computational budgets constrain the spatial and temporal extent of a single generation, and direct video synthesis continues to face challenges in maintaining coherence and dependencies over longer intervals~\cite{videodiffusionsurvey}. Programs and simulation offer a complementary approach to video creation: reusable assets and explicit story orchestration define the content, and execution produces the frames~\cite{code2video,mover}. With fixed assets, programs, and execution configuration, deterministic playback makes the effects of production decisions available for inspection and revision. In a long animation, these effects also shape the story. A character who acquires a key carries it into later interactions, while information conveyed in dialogue can motivate a subsequent reaction. The camera and timing make these developments visible to the audience. Coordinating world state, character behavior, dialogue, and framing across an entire animation therefore requires understanding what each scene actually enacts. When all scenes are generated together, a missed handoff or an unfinished interaction can leave later scenes dependent on an event that the audience never sees.

Temporal conditioning and streaming extend video generation through visual context at multiple timescales, persistent memory, and training for longer autoregressive rollouts~\cite{streamingt2v,longlive,lol}. Planning across scenes organizes recurring entities, scene descriptions, and spatial layouts~\cite{videodirectorgpt}. Executable generation exposes the actions and timing behind a sequence. Code2Video combines outlines and storyboards with code synthesis, scoped refinement, and visual feedback~\cite{code2video}, while spatiotemporal verification and engine feedback guide iterative animation generation~\cite{mover,storyblender}. Sustaining a narrative across scenes also requires the realized outcome of each scene to guide the next. Execution establishes which interactions occur, where characters end up, and what the audience can observe. Revising a scene can change these outcomes and the assumptions available to later scenes. This dependence calls for a representation linking spatial and narrative state, together with a repair process that revises the relevant actions or story beats while retaining satisfactory content.

We propose \method{}, a framework for long horizon animation generation with two modules: \emph{iterative enactment} and \emph{engine reorchestration}. Iterative enactment generates, executes, and inspects each scene from accepted spatial and narrative states. Feedback locates actions or story beats for revision, and local repair retains the surrounding performance. Each accepted scene advances the world and narrative states, allowing the draft to grow from realized outcomes. Engine reorchestration carries this draft into production. Its story, events, geometry, and shared clock guide the mapping of assets, actions, camera cues, and sound to the selected engine. Engine observations guide presentation revisions, and inspection of the assembled animation addresses continuity and timing across scenes. Final rendering replays the frozen production with the accepted story and temporal organization.

Our contributions are threefold: (1) an iterative enactment algorithm based on spatial state transitions that conditions scene generation on accepted world and narrative outcomes and uses execution feedback for local repair; (2) a harness for long video generation that connects draft enactment, engine reorchestration, sequence inspection, and frozen replay; and (3) three generated animation videos that demonstrate the effectiveness of the method.

\begin{figure*}[t]
\centering
\includegraphics[width=\textwidth]{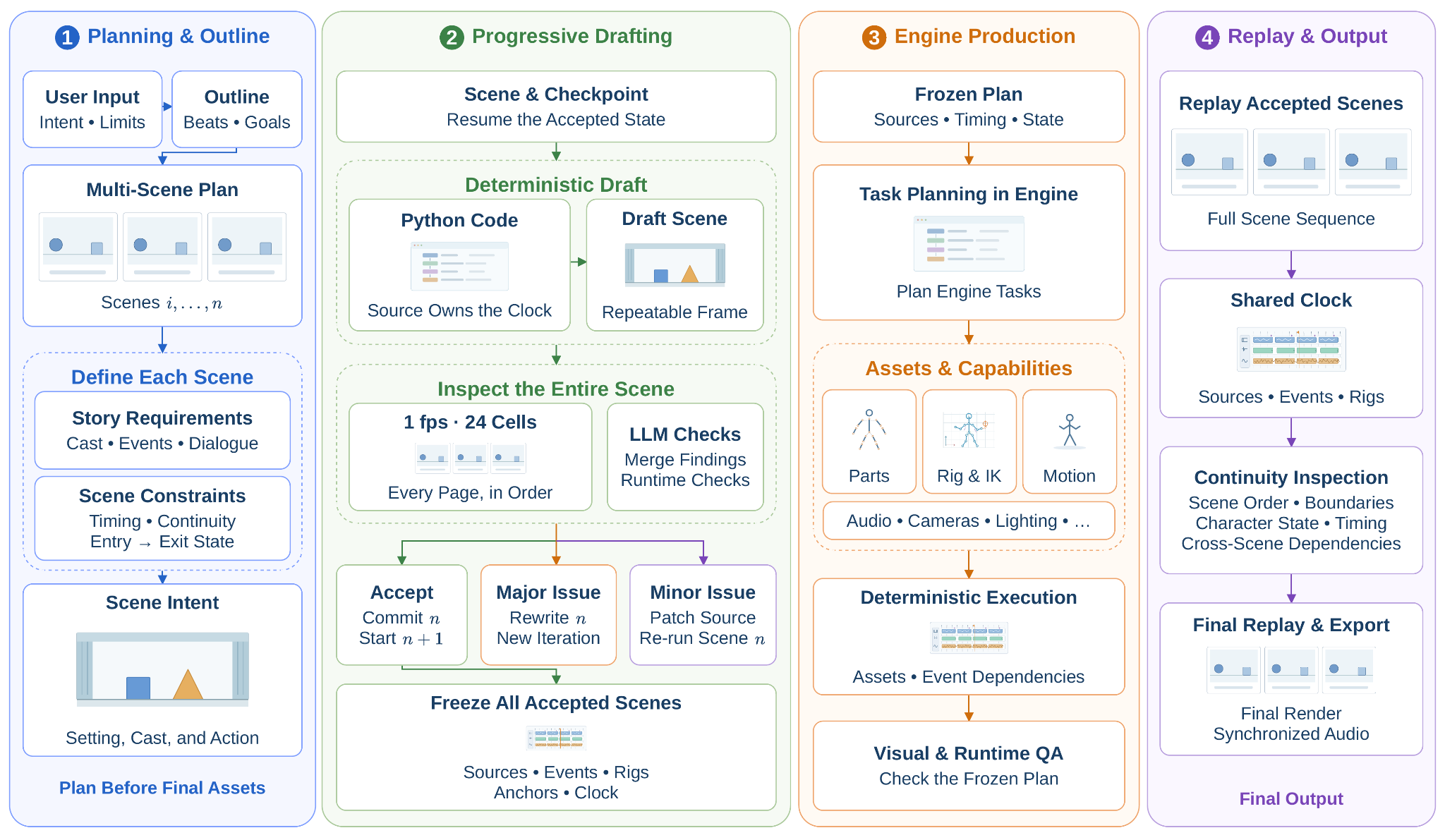}
\caption{Overview of \method{}. Story decomposition supplies the scene plans for iterative enactment. Engine reorchestration carries the accepted draft into production and final replay.}
\label{fig:overview}
\end{figure*}

\section{Related Work}
Seedance~1.0 jointly models video generation from text and images and supports sequences with multiple shots~\cite{seedance1}. Wan combines diffusion transformers with a spatiotemporal variational autoencoder and scalable training for video generation conditioned on text or images~\cite{wan}. HunyuanVideo scales an open video foundation model through data curation, architecture design, and progressive training~\cite{hunyuanvideo}. StreamingT2V uses conditioning at short and long timescales to autoregressively extend videos while maintaining transitions and appearance~\cite{streamingt2v}. LongLive combines causal generation at the frame level, key and value recaching, and streaming long tuning for interactive synthesis of long videos~\cite{longlive}. LoL introduces rotary position embedding jitter across attention heads to mitigate attention sink collapse during long horizon streaming generation~\cite{lol}.

VideoDirectorGPT plans entities, layouts, and consistency information across scenes to guide video synthesis~\cite{videodirectorgpt}. Code2Video combines outlines and storyboards with executable code synthesis, scoped refinement, and visual feedback for educational video production~\cite{code2video}. GPT4Motion generates Blender scripts for physical motion simulation and uses the resulting structural information to condition video generation with diffusion models~\cite{gpt4motion}. MoVer expresses spatiotemporal properties of motion graphics in first order logic and returns failed predicates to a language model for iterative repair~\cite{mover}. StoryBlender combines a continuity memory graph, a unified coordinate space, and feedback verified by the engine to construct and refine 3D storyboards with consistency across shots~\cite{storyblender}.

\section{Method}
\method{} combines story decomposition with iterative enactment and engine reorchestration (Fig.~\ref{fig:overview}). Given a user goal, character descriptions, and production requirements, the planner develops an outline across scenes and a timed script for the current scene. The framework executes and inspects that scene on a shared spatial stage, using the resulting feedback to guide revision. Accepted scenes form a draft whose story, spatial relationships, and realized timing guide asset preparation and engine production. Engine reorchestration presents this performance using the available visual and audio capabilities.

\subsection{Story Decomposition and Script Generation}
The planner extracts the narrative intent from the user goal and develops a structured outline. Each scene specifies its setting, time context, participants, and narrative development. Character descriptions and dependencies across scenes provide persistent context: an acquired object or information revealed in dialogue affects later scenes. The outline establishes the story's direction, and the executable details of each scene are developed after its predecessor has been accepted.

For the current scene, the planner produces timed story beats. Each beat combines a screenplay description with its participants, spoken text, intended result, timing cues, and dependencies on other beats. This representation connects narrative events, such as approaching a companion or handing over a key, to actions that can be executed and inspected. The structured plan is the source for the readable screenplay and subsequent action generation. Descriptions and asset requirements can also specify music and sound effects, which are bound to concrete audio cues during production.

Timing cues establish the intended pacing. Execution refines this schedule using measured speech durations and action completion events, so a response can follow the actual end of a line or interaction. The plan thus connects narrative meaning to the temporal relationships that communicate it. Developing one scene at a time allows its script to incorporate the spatial and narrative outcomes of earlier enactments.

\subsection{Iterative Draft Enactment}
\begin{figure*}[t]
\centering
\includegraphics[width=\textwidth]{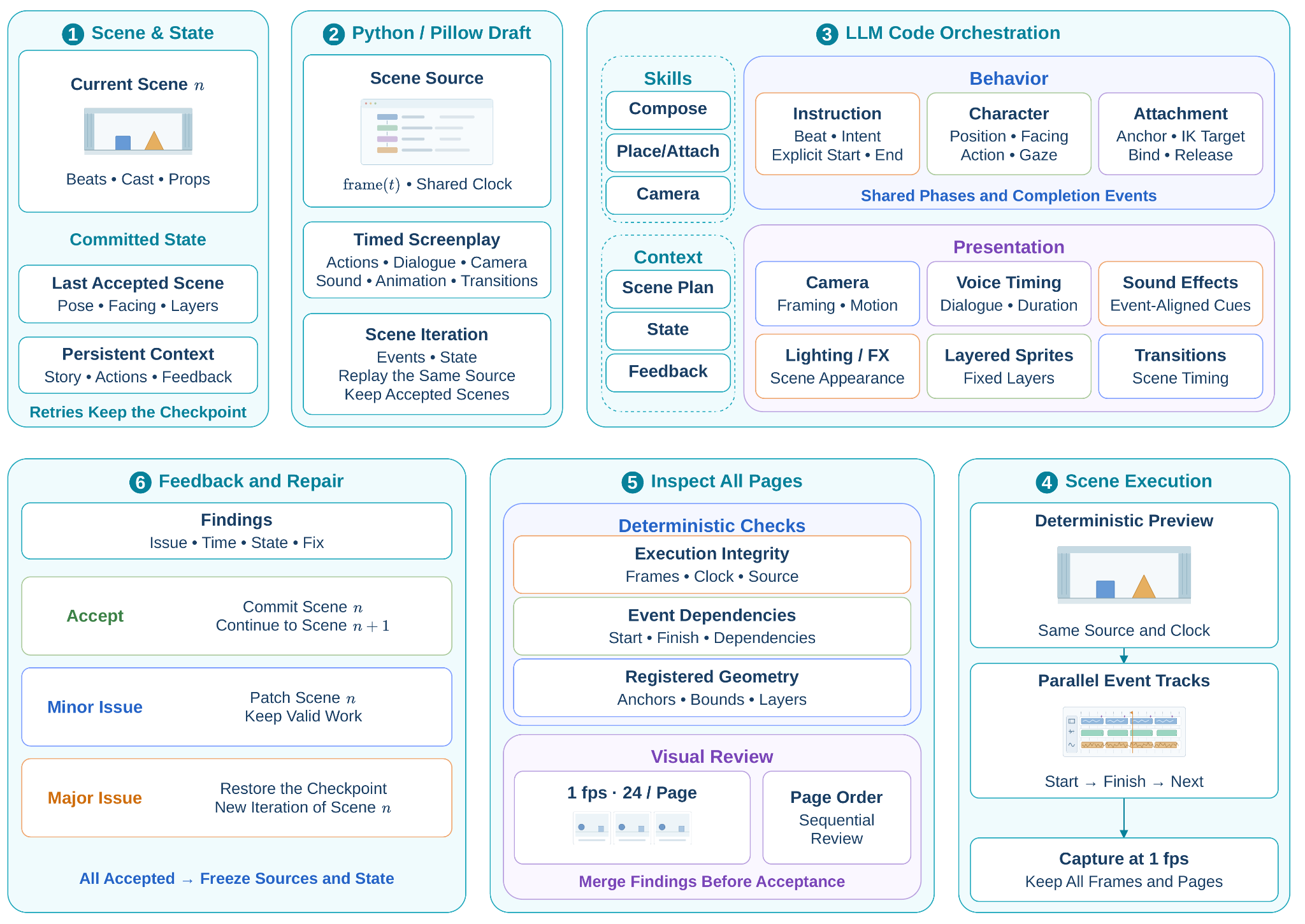}
\caption{Iterative enactment in \method{}. Runtime and visual feedback guide scene acceptance, local repair, and broader revision. Accepted world and narrative states guide the next scene.}
\label{fig:iteration}
\end{figure*}

The draft loop turns each scene plan into a spatially grounded performance (Fig.~\ref{fig:iteration}). At scene boundary $i$, the accepted state is $\state_i=(W_i,H_i)$. The world state $W_i$ records positions, facing directions, visibility, rig poses, attachments, ongoing scene behavior, and the shared clock. The narrative state $H_i$ records established events, character knowledge, and unresolved story threads. Together, they supply the starting conditions and narrative context for the next scene. Continuous scenes inherit the accepted configuration. A declared transition establishes a new location or time while carrying forward the relevant narrative history.

The framework registers stage geometry, character parts, anchors, and supported attachments before generating actions. The scene program schedules movement, gaze, contact, dialogue, and camera behavior on the shared clock. Registered geometry and attachment states ground a handoff, while camera cues frame the resulting interaction. Execution takes place in a lightweight draft environment implemented in Python with Pillow. For revision $r$ of scene $i$, execution gives
\begin{equation}
(\widetilde{W}_i^{(r)},\evidence_i^{(r)})
=\operatorname{Execute}(\program_i^{(r)},W_{i-1}),
\label{eq:execute}
\end{equation}
where $\program_i^{(r)}$ is the candidate program, $\widetilde{W}_i^{(r)}$ its terminal world state, and $\evidence_i^{(r)}$ the runtime events, motion observations, and rendered frames. Each candidate executes from a copy of the same accepted predecessor, providing a common starting state for scene revision.

Inspection combines runtime evidence with visual review in temporal order. The default filmstrip samples one frame per second and groups up to 24 frames per page, including the final partial page. A vision language model reads every page in order against the scene plan and prior narrative state. Runtime evidence describes timing, dependencies, and geometric relationships. Visual review examines the legibility of actions, camera coverage, and character reactions in their narrative context. Denser frame samples and motion observations support inspection of brief contacts and transitions. Findings link each issue to its time, entities, and story beat so that revision can target the relevant content.

Inspection leads to three possible outcomes. \emph{Accept} records the terminal world state and the narrative developments supported by the enactment, then advances to the next scene. \emph{Patch} revises the relevant actions, camera or sound cues, or a bounded set of story beats within the current iteration. It preserves unaffected content and reuses unchanged dialogue audio. \emph{Redo} begins a new iteration of the current scene from the accepted predecessor when broader revision is needed. Both repair routes execute and inspect the revised scene before advancing. For an accepted revision $r^*$, the state transition is
\begin{equation}
\state_i=
(\widetilde{W}_i^{(r^*)},\widetilde{H}_i^{(r^*)}),
\label{eq:commit}
\end{equation}
where $\widetilde{H}_i^{(r^*)}$ contains the narrative developments supported by the accepted enactment. Acceptance requires successful execution and realization of the intended story development. Minor presentation issues can remain when that development is clear. Resource limits and progress criteria bound the iteration, with the latest accepted state retained at termination. The draft grows through these inspected state transitions, grounding later scenes in enacted events.

\subsection{Engine Reorchestration}
Engine reorchestration begins with the accepted draft and original character descriptions. The draft establishes the story, events, source geometry, rigs, attachments, facing directions, and realized timing of motion, dialogue, and camera behavior. This reference performance guides asset preparation and presentation choices. Production artwork supplies appearance, while camera, lighting, and audio shape how the established events are presented.

Character preparation connects draft geometry to production artwork. The framework assembles a transparent reference image on the original draft canvas and combines it with the character description, part structure, rig, and attachments to guide image editing. If the image model requires a different output size, transparent padding followed by cropping to the source rectangle retains the canvas scale and coordinates. Part decomposition and registration establish the geometry of the resulting artwork. Preparation on the original canvas uses polygons to separate visible parts and registers their joints, anchors, layer order, and rig bindings. Designated donor regions and masks supply hidden connections. Generating artwork within the frozen source part canvases retains the corresponding rig, anchor, layer, and facing mappings.

Production binding associates the registered geometry of each part with its draft node and motion. Coordinate mappings between the world stage and the local rig carry actions into the prepared character while retaining the relationships used for contact and attachment. Binding requires corresponding parts and compatible facing configurations. Geometric registration and inspection establish whether the prepared character supports the intended motion. The performance is then baked into editable engine assets, including prefabs, animation clips, and timelines. These assets replay the accepted trajectories and timing.

The available engine capabilities and the narrative events guide presentation. The planner aligns camera offsets, zoom, lighting, and visual effects with events while retaining character trajectories, contacts, dialogue, and event times. Each choice serves a narrative purpose, such as showing a transfer or emphasizing a reaction, and is inspected in the resulting engine sequence.

Audio preparation follows the same event structure. During import, a language model proposes uses, playback modes, tags, and descriptions from audio clip names. Users confirm these annotations before they enter the searchable index. The director queries the index by intended use, style, and description, then considers the selected resource alongside the relevant event and existing audio. Each cue specifies its resource, role, event interval, source offset, gain, and fades. Clip duration and resource type constrain segment selection and repeated playback, while the established voice track remains fixed. This process turns script requirements for music and sound effects into timed production assets.

Engine execution and inspection of the assembled sequence examine visible interactions, continuity across scenes, camera coverage, and audio synchronization. Presentation revisions retain the accepted story and terminal scene states. Changes to facts used by later scenes propagate through the dependent scene chain. Once the production is accepted, the framework freezes asset bindings, motion, and timing. Final rendering replays this performance at the selected quality, carrying the inspected spatial and temporal organization into the output video.

\begin{figure*}[t]
\centering
\includegraphics[width=\textwidth]{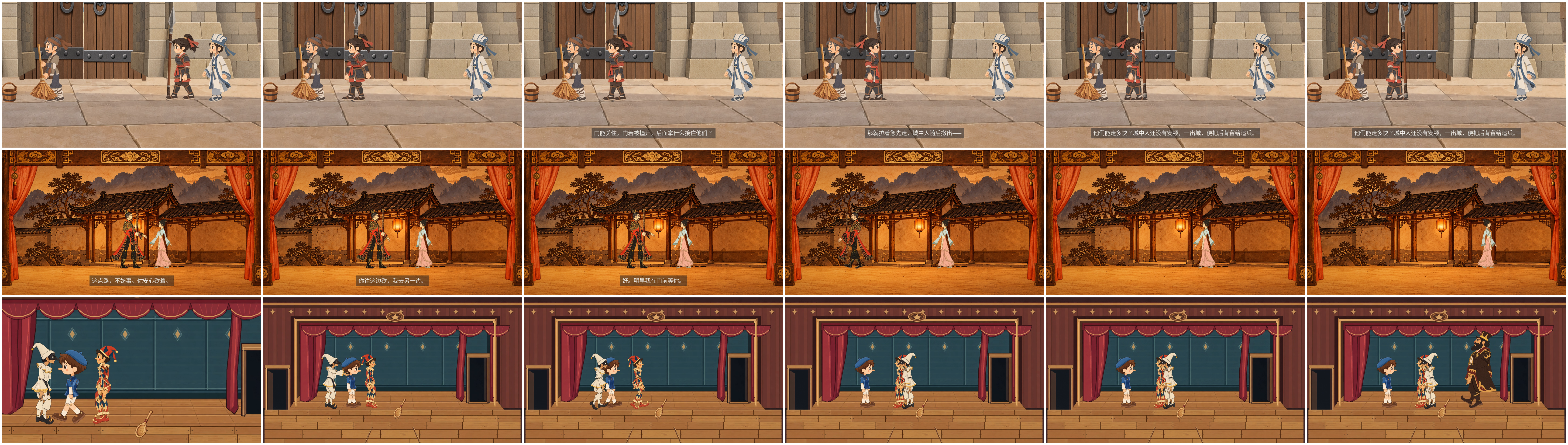}
\caption{Sequences from the three complete animations. Rows show \emph{The Empty City Stratagem} (37 to 62\,s), \emph{Sending Jingniang Home} (312 to 337\,s), and \emph{Pinocchio} (424 to 449\,s), from top to bottom. Each row proceeds from left to right at intervals of five seconds.}
\label{fig:showcases}
\end{figure*}

\begin{figure*}[t]
\centering
\includegraphics[width=\textwidth]{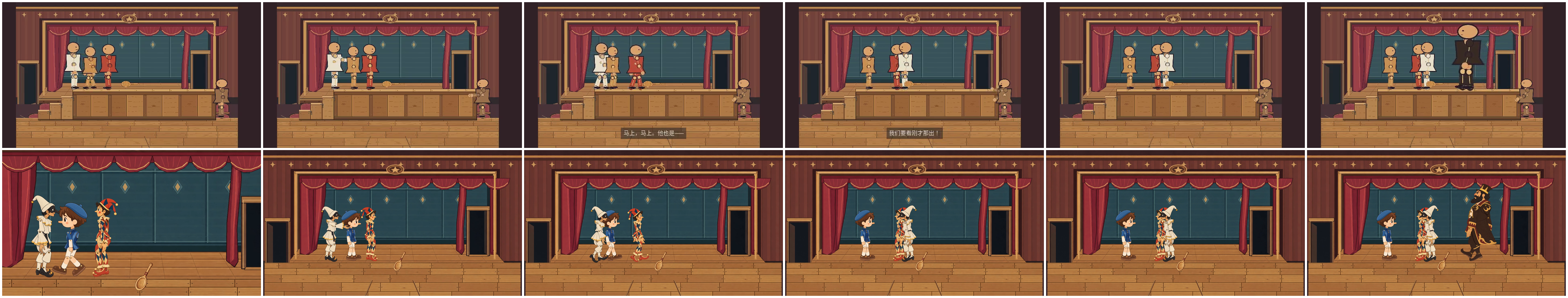}
\caption{Two stages of \emph{Pinocchio} at matched timestamps: draft enactment (top) and production (bottom), sampled every five seconds from 424 to 449\,s. Matched frames show how the draft performance is realized with production artwork and framing.}
\label{fig:pinocchio-stages}
\end{figure*}

\section{Examples}
\label{sec:examples}
\subsection{Setup}
We demonstrate \method{} through three complete animations: \emph{Sending Jingniang Home}, \emph{Pinocchio}, and \emph{The Empty City Stratagem}. Sending Jingniang Home and Pinocchio contain eight and five narrative acts, respectively. The Empty City Stratagem develops two narrative acts across three executable scenes: the inner courtyard, the elevated instrument platform, and the road outside the gate. Each work combines dialogue, character movement, and staged interactions. Each video shows how scene programs organize a narrative into enacted events and how the resulting draft guides engine production.

Figure~\ref{fig:showcases} presents six frames at intervals of five seconds from each animation. Figure~\ref{fig:pinocchio-stages} pairs draft and production frames from \emph{Pinocchio} at matching moments from 424 to 449 seconds, showing the change in visual representation along a shared timeline. The independent \emph{Shadow Duel} draft provides a ten second example of articulated action.

\subsection{Results and Analysis}
Narrative continuity across scenes is expressed through persistent objects, character knowledge, and recurring actions. The three animations span approximately 9.9 to 14.3 minutes, and the sequences in Fig.~\ref{fig:showcases} form local developments within their longer story arcs. In \emph{Pinocchio}, the schoolbook connects the father's sacrifice, the child's decision to attend the theatre, and his later concern about returning home. Selling it provides admission money, while its absence gives meaning to the final responsibility of carrying five coins to his father. \emph{The Empty City Stratagem} connects preparation to interpretation across two acts and three executable scenes. Opening the gates, clearing the entrance, and resuming ordinary sweeping establish the conditions subsequently observed by Sima Yi and Sima Zhao. Their competing interpretations, investigation, and withdrawal follow from those conditions, and a report communicates the outcome to the defenders. \emph{Sending Jingniang Home} sustains the escort relationship through a journey, confession, homecoming, and loss. Looking back, waiting, and making room initially express practical care. The same gestures acquire a different meaning when the final solitary journey returns to an empty road. These connections show how retained narrative context gives later actions significance across changes of location, time, and participants.

\begin{figure}[t]
\centering
\includegraphics[width=\linewidth]{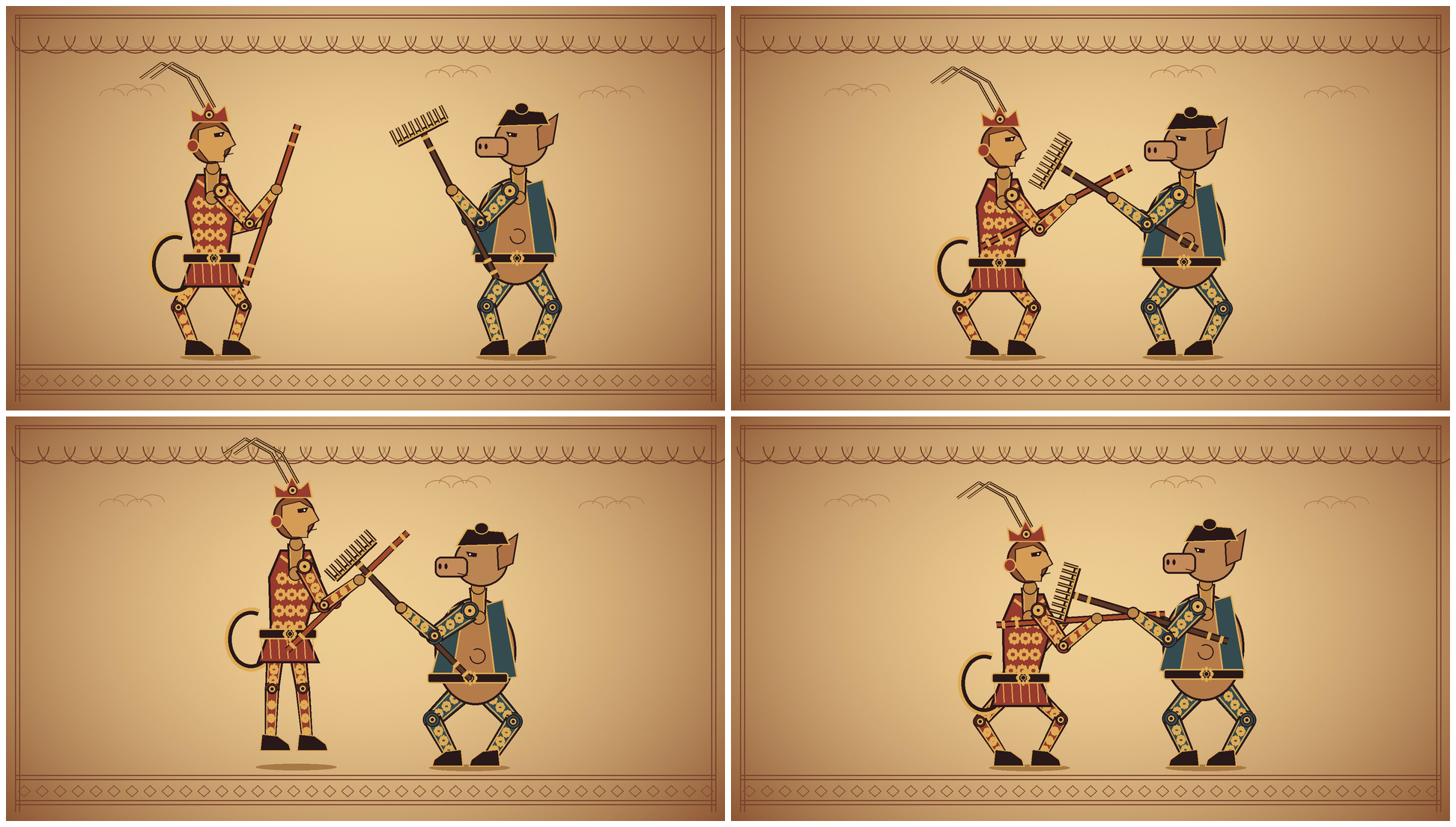}
\caption{An independent shadow puppet action draft lasting ten seconds. Four frames at 0, 2, 4, and 6\,s, read left to right and then top to bottom, show changes in body and weapon configurations.}
\label{fig:shadow-duel}
\end{figure}

Articulated action and spatial orchestration turn these narrative dependencies into visible interactions. At the beginning of \emph{Sending Jingniang Home}, an offscreen call leads through listening and movement to discovery of the captive. Opening the sliding cage panel and stepping aside establish her route out. The mountain encounter varies position, overhead and oblique attacks, staff defence, and counterattack, while Jingniang's retreat and return connect the combat to the interrupted escort. Later, relative positions beneath the eaves express an offer of shelter, and approaches, turns, and gestures accompany changes in the relationship. In \emph{Pinocchio}, a brief shared hold precedes the father's release of the book. Inspection, payment, release, and ticket acquisition then develop the street transaction. Performers gathering around the child interrupt their play, while his later return from the exit toward a threatened companion expresses a change of intention. In \emph{The Empty City Stratagem}, hinged gates expose a passage, the broom moves with its holder, and the weapon rests against a supporting surface. A standing approach leads into seated performance on the elevated platform, with camera changes connecting that position to the gate and approaching opponents. The \emph{Shadow Duel} draft further isolates coordinated body and prop motion (Fig.~\ref{fig:shadow-duel}): frames at 0, 2, 4, and 6 seconds move from separated poses to crossed weapons and a raised body pose. Across these examples, explicit geometry, attachment, and timing organize contact, ownership changes, movement, and group participation within a shared stage.

Speech, mouth motion, subtitles, and music share the temporal organization of the performance. In \emph{Pinocchio} and \emph{The Empty City Stratagem}, recorded character and audio bindings associate each dialogue cue with a speaking entity and registered mouth part. Mouth opening and closing follow activity in the actual speech waveform, with silent intervals returning the mouth to its closed state. Gestures and listener reactions can occur alongside speech, while measured voice intervals supply timing for subtitles and subsequent actions. This coordination connects the visible speaker to the corresponding line in the exchanges between Sima Yi and Sima Zhao and among the puppets. Music also contributes to motivation and dramatic development. In \emph{Pinocchio}, plucked phrases and sustained strings accompany anticipation and the gift of the book; theatre music then motivates the child's change of direction. Changes in register and pacing accompany hesitation over the book, danger near the fire, the return to protect a companion, and farewell. Musical pauses leave room for the explanation about the father's coat and the decision after the child's offer of sacrifice. A metal sound at the placement of the fire tongs is anchored to the completion of that action. In \emph{The Empty City Stratagem}, small alternating arm movements accompany short musical phrases, and the resulting apparent calm becomes part of what the opponents interpret. Sound therefore participates in both event realization and narrative progression.

The separation of enactment and engine production supports changes in appearance and presentation around an established performance. The paired \emph{Pinocchio} frames in Fig.~\ref{fig:pinocchio-stages} depict the gathering of the group, changes in its arrangement, and the later appearance of the theatre master at matched timestamps. Production supplies illustrated characters, detailed costumes, and revised framing around the draft's progression. The common timeline relates these visual changes to the same organized actions and dialogue. The three complete works further present different visual treatments through a shared organization of scenes, event timing, character state, and audio. Together with the compact articulated interaction in \emph{Shadow Duel}, they demonstrate how local action programs can be assembled into extended narratives and carried through to engine output.

\section{Conclusion}
We presented \method{}, a progressive scene programming framework that connects iterative enactment with engine reorchestration. Accepted world and narrative states guide each scene, while runtime and visual feedback locate actions and story beats for revision. The resulting draft supplies the events, spatial relationships, and timing for engine production and frozen replay. Three complete animations demonstrate this approach to constructing extended narratives from executable scenes.

\bibliographystyle{IEEEtran}
\bibliography{references}

@inproceedings{code2video,
  title = {{Code2Video}: A Code-centric Paradigm for Educational Video Creation},
  author = {Chen, Yanzhe and Lin, Kevin Qinghong and Shou, Mike Zheng},
  booktitle = {Proceedings of the 43rd International Conference on Machine Learning},
  series = {Proceedings of Machine Learning Research},
  volume = {306},
  pages = {15370--15396},
  publisher = {PMLR},
  year = {2026},
  month = jul,
  url = {https://proceedings.mlr.press/v306/chen26bx.html}
}

@inproceedings{streamingt2v,
  title = {{StreamingT2V}: Consistent, Dynamic, and Extendable Long Video Generation from Text},
  author = {Henschel, Roberto and Khachatryan, Levon and Poghosyan, Hayk and Hayrapetyan, Daniil and Tadevosyan, Vahram and Wang, Zhangyang and Navasardyan, Shant and Shi, Humphrey},
  booktitle = {Proceedings of the IEEE/CVF Conference on Computer Vision and Pattern Recognition (CVPR)},
  year = {2025},
  month = jun,
  pages = {2568--2577},
  url = {https://openaccess.thecvf.com/content/CVPR2025/html/Henschel_StreamingT2V_Consistent_Dynamic_and_Extendable_Long_Video_Generation_from_Text_CVPR_2025_paper.html}
}

@inproceedings{longlive,
  title={Longlive: Real-time interactive long video generation},
  author={Yang, Shuai and Huang, Wei and Chu, Ruihang and Xiao, Yicheng and Zhao, Yuyang and Wang, Xianbang and Li, Muyang and Xie, Enze and Chen, Yingcong and Lu, Yao and others},
  journal={arXiv preprint arXiv:2509.22622},
  year={2025}
}

@inproceedings{lol,
  title = {{LoL}: Longer than Longer, Scaling Video Generation to Hour},
  author = {Cui, Justin and Wu, Jie and Li, Ming and Yang, Tao and Li, Xiaojie and Wang, Rui and Bai, Andrew and Ban, Yuanhao and Hsieh, Cho-Jui},
  booktitle = {Proceedings of the IEEE/CVF Conference on Computer Vision and Pattern Recognition (CVPR)},
  year = {2026},
  month = jun,
  pages = {38132--38142},
  url = {https://openaccess.thecvf.com/content/CVPR2026/html/Cui_LoL_Longer_than_Longer_Scaling_Video_Generation_to_Hour_CVPR_2026_paper.html}
}

@inproceedings{videodirectorgpt,
  title = {{VideoDirectorGPT}: Consistent Multi-Scene Video Generation via {LLM}-Guided Planning},
  author = {Lin, Han and Zala, Abhay and Cho, Jaemin and Bansal, Mohit},
  booktitle = {Conference on Language Modeling (COLM)},
  year = {2024},
  url = {https://openreview.net/forum?id=sKNIjS2brr}
}

@article{videodiffusionsurvey,
  title = {Video Diffusion Models: A Survey},
  author = {Melnik, Andrew and Ljubljanac, Michal and Lu, Cong and Yan, Qi and Ren, Weiming and Ritter, Helge},
  journal = {Transactions on Machine Learning Research},
  year = {2024},
  eprint = {2405.03150},
  archivePrefix = {arXiv},
  url = {https://openreview.net/forum?id=rJSHjhEYJx}
}

@article{mover,
  title = {{MoVer}: Motion Verification for Motion Graphics Animations},
  author = {Ma, Jiaju and Agrawala, Maneesh},
  journal = {ACM Transactions on Graphics},
  year = {2025},
  volume = {44},
  number = {4},
  articleno = {33},
  numpages = {17},
  doi = {10.1145/3731209},
  url = {https://doi.org/10.1145/3731209},
  pages = {33:1--33:17}
}

@inproceedings{storyblender,
  title = {{StoryBlender}: Inter-shot Consistent and Editable {3D} Storyboard with Spatial-Temporal Dynamics},
  author = {Li, Bingliang and Sun, Zhenhong and Bian, Jiaming and Wu, Yuehao and Wang, Yifu and Li, Hongdong and Bian, Yatao and Mo, Huadong and Dong, Daoyi},
  booktitle = {Computer Vision -- ECCV 2026},
  series = {Lecture Notes in Computer Science},
  volume = {17023},
  pages = {130--147},
  publisher = {Springer Nature Switzerland},
  address = {Cham},
  year = {2026},
  doi = {10.1007/978-3-032-37592-6_8},
  url = {https://doi.org/10.1007/978-3-032-37592-6_8}
}

@article{seedance1,
  title={{Seedance 1.0}: Exploring the Boundaries of Video Generation Models},
  author={Yu Gao and Haoyuan Guo and Tuyen Hoang and Weilin Huang and Lu Jiang and Fangyuan Kong and Huixia Li and Jiashi Li and Liang Li and Xiaojie Li and Xunsong Li and Yifu Li and Shanchuan Lin and Zhijie Lin and Jiawei Liu and Shu Liu and Xiaonan Nie and Zhiwu Qing and Yuxi Ren and Li Sun and Zhi Tian and Rui Wang and Sen Wang and Guoqiang Wei and Guohong Wu and Jie Wu and Ruiqi Xia and Fei Xiao and Xuefeng Xiao and Jiangqiao Yan and Ceyuan Yang and Jianchao Yang and Runkai Yang and Tao Yang and Yihang Yang and Zilyu Ye and Xuejiao Zeng and Yan Zeng and Heng Zhang and Yang Zhao and Xiaozheng Zheng and Peihao Zhu and Jiaxin Zou and Feilong Zuo},
  journal={arXiv preprint arXiv:2506.09113},
  year={2025},
  url={https://arxiv.org/abs/2506.09113},
  eprint = {2506.09113},
  archivePrefix = {arXiv},
  primaryClass = {cs.CV},
  doi = {10.48550/arXiv.2506.09113}
}

@article{wan,
  title={{Wan}: Open and Advanced Large-Scale Video Generative Models},
  author={{Team Wan} and Ang Wang and Baole Ai and Bin Wen and Chaojie Mao and Chen-Wei Xie and Di Chen and Feiwu Yu and Haiming Zhao and Jianxiao Yang and Jianyuan Zeng and Jiayu Wang and Jingfeng Zhang and Jingren Zhou and Jinkai Wang and Jixuan Chen and Kai Zhu and Kang Zhao and Keyu Yan and Lianghua Huang and Mengyang Feng and Ningyi Zhang and Pandeng Li and Pingyu Wu and Ruihang Chu and Ruili Feng and Shiwei Zhang and Siyang Sun and Tao Fang and Tianxing Wang and Tianyi Gui and Tingyu Weng and Tong Shen and Wei Lin and Wei Wang and Wei Wang and Wenmeng Zhou and Wente Wang and Wenting Shen and Wenyuan Yu and Xianzhong Shi and Xiaoming Huang and Xin Xu and Yan Kou and Yangyu Lv and Yifei Li and Yijing Liu and Yiming Wang and Yingya Zhang and Yitong Huang and Yong Li and You Wu and Yu Liu and Yulin Pan and Yun Zheng and Yuntao Hong and Yupeng Shi and Yutong Feng and Zeyinzi Jiang and Zhen Han and Zhi-Fan Wu and Ziyu Liu},
  journal={arXiv preprint arXiv:2503.20314},
  year={2025},
  url={https://arxiv.org/abs/2503.20314},
  eprint = {2503.20314},
  archivePrefix = {arXiv},
  primaryClass = {cs.CV},
  doi = {10.48550/arXiv.2503.20314}
}

@article{hunyuanvideo,
  title={{HunyuanVideo}: A Systematic Framework for Large Video Generative Models},
  author={Weijie Kong and Qi Tian and Zijian Zhang and Rox Min and Zuozhuo Dai and Jin Zhou and Jiangfeng Xiong and Xin Li and Bo Wu and Jianwei Zhang and Kathrina Wu and Qin Lin and Junkun Yuan and Yanxin Long and Aladdin Wang and Andong Wang and Changlin Li and Duojun Huang and Fang Yang and Hao Tan and Hongmei Wang and Jacob Song and Jiawang Bai and Jianbing Wu and Jinbao Xue and Joey Wang and Kai Wang and Mengyang Liu and Pengyu Li and Shuai Li and Weiyan Wang and Wenqing Yu and Xinchi Deng and Yang Li and Yi Chen and Yutao Cui and Yuanbo Peng and Zhentao Yu and Zhiyu He and Zhiyong Xu and Zixiang Zhou and Zunnan Xu and Yangyu Tao and Qinglin Lu and Songtao Liu and Dax Zhou and Hongfa Wang and Yong Yang and Di Wang and Yuhong Liu and Jie Jiang and Caesar Zhong},
  journal={arXiv preprint arXiv:2412.03603},
  year={2024},
  url={https://arxiv.org/abs/2412.03603},
  eprint = {2412.03603},
  archivePrefix = {arXiv},
  primaryClass = {cs.CV},
  doi = {10.48550/arXiv.2412.03603}
}

@inproceedings{gpt4motion,
  title={{GPT4Motion}: Scripting Physical Motions in Text-to-Video Generation via {Blender}-Oriented {GPT} Planning},
  author={Jiaxi Lv and Yi Huang and Mingfu Yan and Jiancheng Huang and Jianzhuang Liu and Yifan Liu and Yafei Wen and Xiaoxin Chen and Shifeng Chen},
  booktitle={Proceedings of the IEEE/CVF Conference on Computer Vision and Pattern Recognition (CVPR) Workshops},
  year={2024},
  pages={1430--1440},
  url={https://openaccess.thecvf.com/content/CVPR2024W/PBDL/html/Lv_GPT4Motion_Scripting_Physical_Motions_in_Text-to-Video_Generation_via_Blender-Oriented_GPT_CVPRW_2024_paper.html},
  doi = {10.1109/CVPRW63382.2024.00150},
  month = jun
}

@IEEEtranBSTCTL{IEEEcontrol,
  CTLuse_forced_etal = {yes},
  CTLmax_names_forced_etal = {6},
  CTLnames_show_etal = {1}
}
\end{document}